\pdfoutput=1
\documentclass[11pt]{article}

\usepackage{ACL2023}

\usepackage{times}
\usepackage{latexsym}

\usepackage[T1]{fontenc}
\usepackage[utf8]{inputenc}

\usepackage{microtype}

\usepackage{inconsolata}

\usepackage{graphicx}
\usepackage{algorithm}
\usepackage[noend]{algpseudocode}

\makeatletter
\renewcommand{\fnum@algorithm}{\fname@algorithm}
\makeatother

\title{Lessons on Parameter Sharing across Layers in Transformers}

\author{Sho Takase\thanks{\ \ A part of this work was done when the author was at Tokyo Institute of Technology.} \hspace{5mm}
  Shun Kiyono \\
  LINE Corporation \\
  \texttt{\{sho.takase, shun.kiyono\}@linecorp.com} \\
  }

\begin{document}
\maketitle
\begin{abstract}
We propose a novel parameter sharing method for Transformers~\cite{NIPS2017_7181}.
The proposed approach relaxes a widely used technique, which shares the parameters of one layer with all layers such as Universal Transformers~\cite{dehghani2019}, to improve the efficiency.
We propose three strategies: \textsc{sequence}, \textsc{cycle}, and \textsc{cycle (rev)} to assign parameters to each layer.
Experimental results show that the proposed strategies are efficient in terms of the parameter size and computational time in the machine translation task.
We also demonstrate that the proposed strategies are effective in the configuration where we use many training data such as the recent WMT competition.
Moreover, we indicate that the proposed strategies are also more efficient than the previous approach~\cite{dehghani2019} on automatic speech recognition and language modeling tasks.
\end{abstract}

\section{Introduction}
\label{sec:intro}
Transformer-based methods have achieved notable performance in various NLP tasks~\cite{NIPS2017_7181,devlin-etal-2019-bert,NEURIPS2020_1457c0d6}.
In particular, \citet{NEURIPS2020_1457c0d6} indicated that the larger parameter size we prepare, the better performance the model achieves.
However, the model which is composed of many parameters occupies a large part of a GPU memory capacity.
Thus, it is important to explore a parameter efficient way, which achieves better performance than a basic model with the same parameter size.

Parameter sharing is a widely used technique as a parameter efficient way~\cite{dehghani2019,Dabre_Fujita_2019,lan2020}.
\citet{dehghani2019} proposed Universal Transformer which consists of parameters for only one layer of a Transformer-based encoder-decoder, and uses these parameters $N$ times for an $N$-layered encoder-decoder.
\citet{Dabre_Fujita_2019} and \citet{lan2020} also used such parameter sharing across layers for their Transformers.

\begin{figure}[!t]
  \centering 
  \includegraphics[width=8cm]{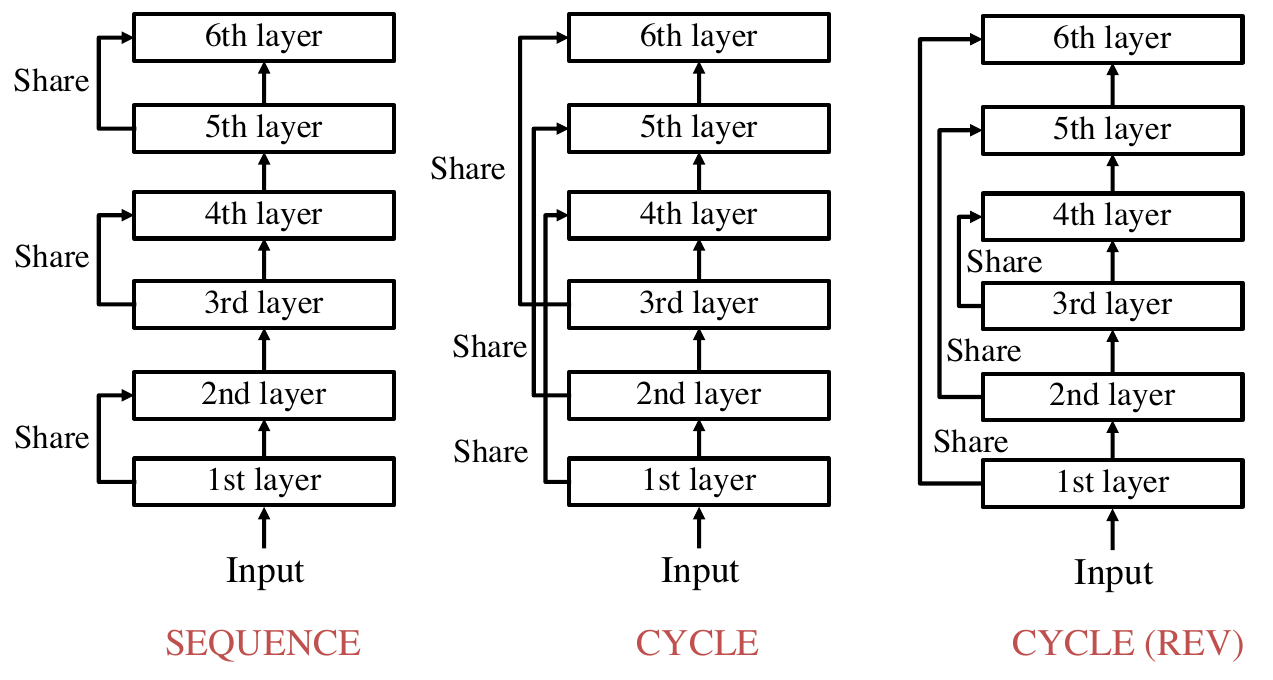}
   \caption{Examples of three parameter assignment strategies proposed in this study when we set $M=3$ and $N=6$.}
   \label{fig:overview}
\end{figure}

\citet{dehghani2019} reported that Universal Transformer achieved better performance than the vanilla Transformer in machine translation if the parameter sizes of both models are (almost) the same.
However, when we prepare the same number of parameters for Universal Transformer and vanilla Transformer, the dimension sizes of each layer in Universal Transformer are much larger than ones in the vanilla Transformer.
Thus, Universal Transformer requires much more computational time since its weight matrices are larger.
For example, Universal Transformer requires twice as much training time as the vanilla Transformer in WMT English-to-German dataset, which is a widely used machine translation dataset (see Table \ref{tab:exp_main_result}).

In this paper, we propose a new parameter sharing method that is faster than using the same parameters for all layers such as Universal Transformers.
Universal Transformers raise their expressiveness power by increasing the size of weight matrices for each layer.
On the other hand, stacking (more) layers is another promising approach to raise expressiveness power of neural methods~\cite{7780459}.
Thus, the most straight-forward way to make Universal Transformers faster is stacking layers with smaller weight matrices for each layer.
However, the approach using the same parameters for all layers limits the improvement of stacking layers~\cite{Dabre_Fujita_2019}.
Therefore, instead of preparing parameters for only one layer, we prepare parameters for $M$ layers to construct an $N$-layered encoder-decoder, where $1 \leq M \leq N$.
In other words, the proposed method relaxes the parameter sharing strategy in previous studies~\cite{dehghani2019,Dabre_Fujita_2019,lan2020}.
Because this relaxation addresses the above limitation of improvement by stacking layers, the proposed method can be fast by stacking layers with using small weight matrices for each layer.
For the actual parameter assignment strategies, we provide several simple examples (Figure~\ref{fig:overview}) and investigate their performance empirically.
The main focus of this study is to demonstrate that such simple strategies can be a better alternative to the existing parameter sharing strategy used in Universal Transformers.

We mainly conduct experiments on machine translation datasets.
Experimental results show that the proposed method achieves slightly better scores to the previous method, that assigns parameters of one layer to all layers, with smaller computational time.
In addition, we indicate that the proposed method outperforms the previous parameter sharing method when we spend almost the same training time.
Moreover, we conduct experiments on automatic speech recognition and language modeling tasks (Section \ref{sec:exp_asr} and Appendix \ref{sec:exp_lm}).
Experimental results on these tasks also indicate that the proposed method are also efficient in these situations.

\section{Proposed Method}
\begin{figure}[!t]
  \begin{algorithm}[H]
  % \small
  \caption{Encoder Construction}
  \begin{algorithmic}[1]
   \renewcommand{\algorithmicrequire}{\textbf{Input:}}
   \renewcommand{\algorithmicensure}{\textbf{Output:}}
   \Require the total number of layers $N$, number of independent layers $M$, sharing strategy \textsc{Type} $\in$ \{\textsc{sequence, cycle, cycle (rev)}\}
   \Ensure $\textrm{enc}_1$, ..., $\textrm{enc}_N$
   \For{$i$ \textrm{in} $[1, ..., N]$}
     \If{$i$ == $1$}
     \State $\textrm{enc}_i \gets \textrm{CreateNewLayer}$
     \ElsIf{\textsc{Type} == \textsc{sequence}}
       \If{$(i - 1) \bmod \lfloor N / M \rfloor == 0$}
       \State $\textrm{enc}_i \gets \textrm{CreateNewLayer}$
       \Else
       \State $\textrm{enc}_i \gets \textrm{enc}_{i-1}$
       \EndIf
     \ElsIf{\textsc{Type} == \textsc{cycle}}
       \If{$i \leq M$}
       \State $\textrm{enc}_i \gets \textrm{CreateNewLayer}$
       \Else
       \State $\textrm{enc}_i \gets \textrm{enc}_{((i - 1) \bmod M) + 1}$
       \EndIf
     \ElsIf{\textsc{Type} == \textsc{cycle (rev)}}
       \If{$i \leq M$}
       \State $\textrm{enc}_i \gets \textrm{CreateNewLayer}$
       \ElsIf{$i \leq (M \times (\lceil N / M \rceil - 1))$}
       \State $\textrm{enc}_i \gets \textrm{enc}_{((i - 1) \bmod M) + 1}$
       \Else
       \State $\textrm{enc}_i \gets \textrm{enc}_{M - ((i - 1) \bmod M)}$       
       \EndIf
     \EndIf
   \EndFor
  % \State $s \leftarrow 0$
  \end{algorithmic}
  \end{algorithm}
  \caption{Proposed parameter assignment strategies for encoder construction. CreateNewLayer is a function that creates a new encoder layer.}
  \label{fig:enc_construction}
\end{figure}

As described in Section \ref{sec:intro}, we use parameters for $M$ layers in the construction of an $N$-layered Transformer-based encoder-decoder.
We provide three examples for the parameter assignment: \textsc{sequence}, \textsc{cycle}, and \textsc{cycle (rev)}.
This section describes these parameter assignment strategies.

Figure \ref{fig:overview} shows examples of three parameter assignment strategies for an encoder side when we set $M=3$ and $N=6$.
Let $\textrm{enc}_i$ be the $i$-th layer of an encoder.
Figure \ref{fig:enc_construction} describes the algorithm to assign each parameter to each layer of the encoder.
For the decoder side, we assign each parameter with the same manner.

\subsection{\textsc{sequence}}
The simplest strategy is to assign the same parameters to sequential $\lfloor N / M \rfloor$ layers.
We name this strategy \textsc{sequence}.
For example, when we set $M=3$ and $N=6$, two sequential layers share their parameters as illustrated in Figure \ref{fig:overview}.

\subsection{\textsc{cycle}}
In \textsc{cycle}, we stack $M$ layers whose parameters are independent from each other.
Then, we repeat stacking the $M$ layers with the identical order to the first $M$ layers until the total number of layers reaches $N$.
When we set $M=3$ and $N=6$, we stack $3$ layers twice as illustrated in Figure \ref{fig:overview}.

\subsection{\textsc{cycle (rev)}}
\citet{liu-etal-2020-understanding} and \citet{takase:2022:b2t} reported that higher decoder layers tends to obtain larger gradient norms\footnote{In particular, this property is observed during warm-up when we use the post layer normalization (Post-LN) setting, which is originally used in \newcite{NIPS2017_7181} and widely used in machine translation.}.
Their report implies that higher layers require more degrees of freedom than lower layers for their expressiveness.
In other words, lower layers probably have redundant parameters compared to higher layers.
Thus, we propose the \textsc{cycle (rev)} strategy reusing parameters of lower layers in higher layers.

In this strategy, we repeat stacking $M$ layers in the same manner as \textsc{cycle} until $M * (\lceil N / M \rceil - 1)$ layers.
For the remaining layers, we stack $M$ layers in the reverse order.
When we set $M=3$ and $N=6$, we stack $3$ layers and then stack the $3$ layers in the reverse order as in Figure \ref{fig:overview}.
% Thus, the lowest layer and highest layer share their parameters.
Thus, the lowest layer and highest layer share parameters.

\section{Experiments on Machine Translation}
\label{sec:exp_mt}

We investigate the efficiency of the proposed parameter sharing strategies. 
In detail, we indicate that our proposed strategies are faster than Universal Transformers while achieving comparable (or better) performance when we use the same parameter size.
In this section, we conduct experiments on machine translation datasets.
First, we focus on the English-to-German translation task because this task is widely used in the previous studies~\cite{NIPS2017_7181,ott-etal-2018-scaling,dehghani2019,kiyono-etal-2020-tohoku}.
We conduct comparisons based on following aspects: (i) comparison with Universal Transformers in terms of efficiency and 
(ii) comparison with models without parameter sharing across layers to investigate whether our proposed strategies can achieve comparable (or better) performance to the models with larger memory footprint.

In addition to the widely used training data, we conduct experiments on a large amount of training dataset in the English-to-German translation task.
Then, we investigate if our findings are consistent in other language direction (i.e., German-to-English) and other language pair (i.e., English-to-French and French-to-English).
We describe details in the following subsections.

\begin{table*}[!t]
  \centering{}
  \footnotesize
  \begin{tabular}{ l | c c | r | r | c c c c c c c | c} \hline
  Method & $M$ & $N$ & \multicolumn{1}{c|}{\#Params} & Speed & 2010 & 2011 & 2012 & 2013 & 2014 & 2015 & 2016 & Avg. \\ \hline \hline
  Vanilla & 6 & 6 & 61M & $\times$2.02 & 24.14 & 21.93 & 22.25 & 26.14 & 27.05 & 29.59 & 34.23 & 26.48 \\
  Universal & 1 & 6 & 63M & $\times$1.00 & 24.37 & 22.33 & 22.70 & 26.40 & 27.65 & 30.24 & 34.60 & 26.90 \\
  Universal (deep) & 1 & 12 & 63M & $\times$0.52 & 24.42 & 22.30 & 22.61 & 26.52 & 27.76 & 29.75 & 34.01 & 26.77 \\
  Universal (small) & 1 & 6 & 24M & $\times$2.52 & 22.89 & 21.11 & 21.29 & 24.75 & 24.71 & 28.16 & 32.81 & 25.10 \\
  \textsc{sequence} & 6 & 12 & 61M & $\times$1.31 & 24.65 & 22.32 & 22.83 & \textbf{26.98} & 27.88 & 30.27 & \textbf{34.99} & \textbf{27.13} \\  
  \textsc{cycle} & 6 & 12 & 61M & $\times$1.31 & 24.51 & 22.43 & 22.69 & 26.61 & \textbf{27.91} & \textbf{30.37} & 34.77 & 27.04 \\
  \textsc{cycle (rev)} & 6 & 12 & 61M & $\times$1.31 & \textbf{24.66} & \textbf{22.47} & \textbf{22.87} & 26.68 & 27.72 & \textbf{30.37} & 34.81 & 27.08 \\ \hline
  \textsc{sequence} & 6 & 18 & 61M & $\times$0.98 & 24.53 & 22.44 & 22.73 & 26.59 & 27.73 & 30.30 & 34.80 & 27.02 \\
  \textsc{cycle} & 6 & 18 & 61M & $\times$0.98 & 24.74 & 22.60 & 23.04 & \textbf{26.89} & \textbf{28.14} & 30.54 & 34.79 & 27.25 \\
  \textsc{cycle (rev)} & 6 & 18 & 61M & $\times$0.98 & \textbf{24.93} & \textbf{22.77} & \textbf{23.09} & 26.88 & 28.09 & \textbf{30.60} & \textbf{34.84} & \textbf{27.31} \\ \hline \hline
  \multicolumn{13}{c}{Methods consisting of a large number of parameters for reference} \\ \hline \hline
  Vanilla (big) & 6 & 6 & 210M & $\times$0.81 & 24.31 & 22.21 & 22.75 & 26.39 & 28.28 & 30.35 & 33.40 & 26.81 \\
  Vanilla (deep) & 18 & 18 & 149M & $\times$0.96 & 24.54 & 22.30 & 22.75 & 26.57 & 28.03 & 30.24 & 34.19 & 26.94 \\ \hline
  \end{tabular}
  \caption{The number of layers, number of parameters, computational speeds based on the Universal configuration, BLEU scores on newstest2010-2016, and averaged scores when we trained each method on widely used WMT 2016 English-to-German training dataset. Scores in bold denote the best results for each set. The results of our proposed strategies are statistically significant ($p < 0.05$) in comparison with Universal. The lowest part indicates results of methods consisting of a large number of parameters for reference.\label{tab:exp_main_result}}
\end{table*}

\subsection{Standard Setting}
\label{sec:exp_standard_setting}

\subsubsection{Datasets}
We used the WMT 2016 training dataset, which is widely used in previous studies~\cite{NIPS2017_7181,ott-etal-2018-scaling,takase-kiyono-2021-rethinking}.
This dataset contains 4.5M English-German sentence pairs.
Following previous studies, we constructed a vocabulary set with BPE~\cite{sennrich-etal-2016-neural} in the same manner.
We set the number of BPE merge operations at 32K and shared the vocabulary between the source and target languages.
We measured case-sensitive detokenized BLEU with SacreBLEU~\cite{post-2018-call}\footnote{The BLEU score computed by SacreBLEU is often lower than the score obtained by the procedure of \newcite{NIPS2017_7181} as reported in \newcite{ott-etal-2018-scaling}. In fact, when we used the same procedure as \newcite{NIPS2017_7181}, \textsc{sequence} of $M=6, N=12$ in Table \ref{tab:exp_main_result} achieved 29.40 in the averaged BLEU score in newstest2014 and the best model in Table \ref{tab:exp_wmt2020_result} achieved 35.14 in the averaged BLEU score in newstest2014. However, since \newcite{post-2018-call} encouraged using SacreBLEU for the compatibility of WMT results, we used SacreBLEU.}.

\subsubsection{Methods}
For the proposed parameter assignment strategies, we fixed $M = 6$ and set $N = 12, 18$ based on the Vanilla configuration below.
We compare the proposed strategies with the following baselines.

\noindent\textbf{Vanilla}: This is the original Transformer (base) setting in \cite{NIPS2017_7181}.
To stabilize the training, we applied Admin~\cite{liu-etal-2020-understanding}.
See Section \ref{sec:relatedwork} for more details of Admin.

\noindent\textbf{Universal}: As the parameter sharing strategy in previous studies such as Universal Transformers~\cite{dehghani2019}, we set $M = 1$\footnote{The original Universal Transformers~\cite{dehghani2019} use the sinusoidal positional encoding for each layer and adaptive computation time technique~\cite{graves2017adaptive} but we omitted them in this study to focus on the difference among parameter sharing strategies.}.
In this setting, we increased the dimensions of each layer for a fair comparison in terms of the number of parameters.
This configuration corresponds to the Universal Transformer base setting in \cite{dehghani2019}.
Moreover, we prepared the model using twice as many layers to investigate the effect of stacking many layers in Universal Transformers.
We call this setting \textbf{Universal (deep)}.
In addition, we prepared \textbf{Universal (small)} whose dimension sizes are the identical to ones of Transformer (base).

Furthermore, we prepare two models that consist of a large number of parameters for reference.

\noindent\textbf{Vanilla (big)}: This is the original Transformer (big) setting in \cite{NIPS2017_7181}.

\noindent\textbf{Vanilla (deep)}: We stacked layers until $N = 18$ in the Vanilla configuration.

\subsubsection{Results}
Table \ref{tab:exp_main_result} shows BLEU scores on newstest2010-2016 for each method.
We trained three models with different random seeds, and reported the averaged scores.
Table \ref{tab:exp_main_result} also shows the total number of parameters and computational speeds\footnote{We regard processed tokens per second during the training as the computational speed.}.
The computational speed is based on the speed of Universal.

\paragraph{(i) Comparison with Universal in terms of efficiency}
In the comparison between Universal and Vanilla, Universal achieved better scores although their parameter sizes are almost the same.
This result is consistent with the report in \cite{dehghani2019}.
However, the training time of Universal is more than twice as much as the one of Vanilla.
In addition, Universal (deep) didn't improve the performance from Universal, and thus stacking many layers have small effect on BLEU scores when the model shares parameters of one layer with all layers.

In contrast, the proposed strategies (\textsc{sequence}, \textsc{cycle}, and \textsc{cycle (rev)}) were faster and achieved slightly better scores than Universal when we set $M = 6$ and $N = 12$.
Thus, our proposed parameter sharing strategies are more efficient than Universal in terms of the parameter size and computational time.

In comparison among Universal (small) and the proposed strategies, Universal (small) was faster\footnote{We used the same dimension sizes for Vanilla and Universal (small) but their training speeds are different from each other. Since Universal (small) consists of small parameters, the computational time for updating is smaller than Vanilla.} but the configuration drastically sacrificed BLEU scores.
These results imply that the strategy in Universal Transformer, which shares parameters of one layer with all layers, damages computational time or the quality of output sequences.
In comparison with those Universal configurations, our proposed strategies improved both of the computational speed and BLEU scores.

\begin{figure}[!t]
  \centering 
  \includegraphics[width=7.5cm]{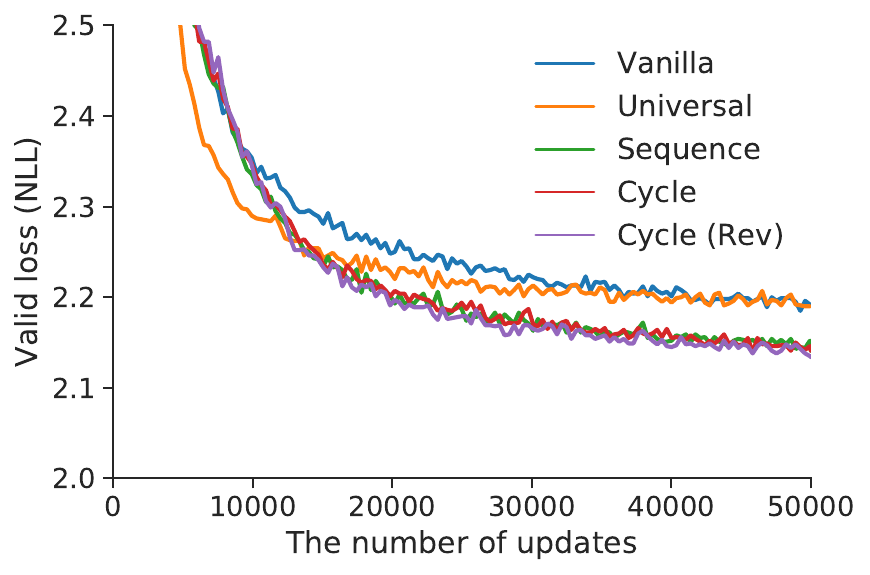}
   \caption{Negative log-likelihood (NLL) of each method on newstest2013. For our proposed parameter sharing strategies, we used $M=6$ and $N=12$.}
   \label{fig:valid_nll_by_time}
\end{figure}

Figure \ref{fig:valid_nll_by_time} illustrates the negative log-likelihood (NLL) values on newstest2013 for each training step.
In this figure, we used $M=6$ and $N=12$ for our proposed strategies.
This figure shows that Universal achieved better NLL values in the beginning of the training but the proposed strategies outperformed others when the training step is larger than 15,000.
When we have finished training, the proposed strategies achieved better NLL values than Universal (and Vanilla).
This result also indicates that the proposed strategies achieved better performance.
We emphasize that the proposed strategies reached this better performance with small computational time in comparison with Universal because the proposed strategies are faster as in Table \ref{tab:exp_main_result}.

\paragraph{(ii) Comparison with models without parameter sharing across layers}
The lowest part of Table \ref{tab:exp_main_result} indicates results when we prepared more parameters.
We trained these models to investigate the performance of models without parameter sharing across layers.
In other words, the purpose of these settings are comparison with models using larger memory footprint.
As shown in Table \ref{tab:exp_main_result}, the proposed strategies achieved better performance than models consisting of a large number of parameters in the averaged BLEU scores of newstest2010-2016.
This result implies that the proposed parameter sharing strategies are not only efficient but also effective in constructing better encoder-decoder models.

\begin{table*}[!t]
  \centering{}
  \footnotesize
  \begin{tabular}{ l | c | c c c c c c c c c | c } \hline
  Method & \#Params & 2010 & 2011 & 2012 & 2013 & 2014 & 2015 & 2016 & 2018 & 2019 & Avg. \\ \hline \hline
  \multicolumn{12}{c}{Genuine training data} \\ \hline \hline
  Vanilla & 242M & 26.53 & 24.09 & 24.51 & 28.51 & 31.40 & 33.52 & 39.08 & 47.11 & 42.80 & 33.06 \\
  Universal & 249M & 27.00 & 24.20 & 24.96 & 28.94 & 31.73 & 33.53 & 39.38 & 47.54 & 43.11 & 33.38 \\
  \textsc{sequence} & 242M & 27.31 & 24.24 & 24.86 & 29.15 & 31.90 & 33.84 & 39.93 & 48.15 & 43.12 & 33.61 \\
  \textsc{cycle} & 242M & 27.23 & 24.45 & 25.13 & 29.12 & \textbf{32.10} & \textbf{34.04} & 39.82 & 48.11 & 43.19 & 33.69 \\
  \textsc{cycle (rev)} & 242M & \textbf{27.37} & \textbf{24.46} & \textbf{25.14} & \textbf{29.16} & 32.06 & 33.98 & \textbf{40.28} & \textbf{48.34} & \textbf{43.43} & \textbf{33.80} \\ \hline \hline
  \multicolumn{12}{c}{+ Synthetic (back-translated) data} \\ \hline \hline
  \citet{kiyono-etal-2020-tohoku} & 514M & - & - & - & - & 33.1 \  & - & - & \textbf{49.6} \  & \textbf{42.7} \ & - \\
  \textsc{cycle (rev)} & 343M & \textbf{28.29} & \textbf{24.99} & \textbf{25.98} & \textbf{30.01} & \textbf{33.54} & \textbf{34.93} & \textbf{41.37} & \textbf{49.55} & 42.18 & \textbf{34.54} \\ \hline
  \end{tabular}
  \caption{BLEU scores on newstest2010-2016, 2018, and 2019. We add newstest2018 and 2019 to the set in the standard setting to compare the top system on WMT 2020 \cite{kiyono-etal-2020-tohoku}.\label{tab:exp_wmt2020_result}}
\end{table*}

\begin{table*}[!t]
  \centering{}
  \footnotesize
  \begin{tabular}{ l | c c | c c | c c | c c } \hline
  & \multicolumn{2}{}{} & \multicolumn{2}{|c|}{German-to-English} & \multicolumn{2}{c|}{English-to-French} & \multicolumn{2}{c}{French-to-English} \\
  Method & $M$ & $N$ & 2013 & 2014 & 2013 & 2014 & 2013 & 2014 \\ \hline \hline
  Vanilla & 6 & 6 & 30.48 & 30.96 & 33.41 & 38.41 & 33.48 & 36.06 \\
  Universal & 1 & 6 & 31.06 & 31.32 & 33.58 & 38.84 & 33.83 & 37.11 \\
  \textsc{sequence} & 6 & 18 & 31.31 & 31.97 & 34.49 & \textbf{40.18} & \textbf{34.26} & 37.45 \\
  \textsc{cycle} & 6 & 18 & \textbf{31.46} & \textbf{32.18} & 34.50 & 40.17 & 33.97 & \textbf{37.59} \\
  \textsc{cycle (rev)} & 6 & 18 & 31.32 & 32.12 & \textbf{34.67} & 40.13 & 34.16 & 37.32 \\ \hline
  \end{tabular}
  \caption{The number of layers and BLEU scores on each dataset. Each method is composed of almost the same number of parameters.\label{tab:exp_other_result}}
\end{table*}

\subsection{High Resource Setting}
\label{sec:exp_high_resource}

\subsubsection{Datasets}
In the high resource setting, we constructed 44.2M translation sentence pairs as a training dataset with the procedures of \cite{kiyono-etal-2020-tohoku} which achieved the best result in the WMT 2020 news translation task.
In addition, we augmented the training data by using the back-translation technique~\cite{sennrich:2016:backtrans} in the same manner as \cite{kiyono-etal-2020-tohoku}.
We obtained 284.3M pairs as synthetic training data.
For evaluation, we add newstest2018 and 2019 to the set used in Section \ref{sec:exp_standard_setting} to because \cite{kiyono-etal-2020-tohoku} used these two test sets.
In the same as Section \ref{sec:exp_standard_setting}, we measured case-sensitive detokenized BLEU with SacreBLEU.

\subsubsection{Methods}
We used the original Transformer (big) setting~\cite{NIPS2017_7181} as our baseline in using genuine training data.
We call this setting \textbf{Vanilla} in this experiment.
Moreover, we also prepared \textbf{Universal}, which shares the parameters with all layers, namely, $M=1, N=6$.
We increased the dimensions of each layer in Universal to make their parameter size almost the same as others.
For the proposed strategies, we used $M = 6$ and $N = 12$.

In using both of the genuine and synthetic (back-translated) datasets, we applied \textsc{cycle (rev)} to the \textsc{Base} setting in \cite{kiyono-etal-2020-tohoku} because \textsc{cycle (rev)} achieved the best BLEU scores on most test sets in Table \ref{tab:exp_main_result}.
We also used $M = 6$ and $N = 12$ in this configuration.
We compare the reported scores of the best model in \cite{kiyono-etal-2020-tohoku}.
Their model is composed of 9 layers (i.e., $M = 9$ and $N = 9$); thus, it contains considerably more parameters than ours.

\subsubsection{Results}
Table \ref{tab:exp_wmt2020_result} shows BLEU scores of each method on each test set.
Similar to the experiments in Section \ref{sec:exp_standard_setting}, we reported the averaged scores of three models trained with different random seeds.
Table \ref{tab:exp_wmt2020_result} also shows the total number of parameters\footnote{The parameter sizes of Vanilla (big) in Table \ref{tab:exp_main_result} and Vanilla in Table \ref{tab:exp_wmt2020_result} are different from each other due to the difference of sharing embeddings. Following \cite{kiyono-etal-2020-tohoku}, we did not share embeddings in the high resource setting.}.

Table \ref{tab:exp_wmt2020_result} shows that the proposed strategies achieved better BLEU scores than Vanilla and Universal when we prepared almost the same number of parameters.
This result indicates that the proposed strategies are also parameter efficient in the high resource setting.
In addition, since we used $M = 6$ and $N = 12$ for proposed strategies, they are also more efficient than Universal in terms of computational time (see Table \ref{tab:exp_main_result}).

When we used additional synthetic data for training in the same manner as \cite{kiyono-etal-2020-tohoku}, \textsc{cycle (rev)} achieved comparable BLEU scores to the best system of \cite{kiyono-etal-2020-tohoku} except for newstest2019\footnote{For newstest2019, synthetic data might harm the quality of a model because models trained with only genuine data outperformed those trained with both data.} even though the parameter size of \textsc{cycle (rev)} was smaller than theirs.
This result indicates that \textsc{cycle (rev)} is also efficient in the construction of models for recent competitive tasks.
In addition, this result implies that our proposed strategies can be used in the configuration where we train many parameters with a tremendous amount of data such as recent pre-trained language models, e.g., GPT series~\cite{NEURIPS2020_1457c0d6}.
We investigate the effect of the proposed strategies on language models in Appendix \ref{sec:exp_lm}.

\subsection{Other Direction and Language Pair}
\label{sec:exp_other_configurations}

\subsubsection{Datasets}
We conduct experiments on the other direction and language pair.
For the German-to-English training dataset, we used the identical data in Section \ref{sec:exp_standard_setting}.
For English-to-French and French-to-English, we used the WMT 2014 training dataset.
We applied the same pre-processing as in \cite{ott-etal-2018-scaling}, and used 35.8M English-French sentence pairs.
Each configuration, we used newstest2013 and newstest2014 as valid and test sets, respectively.
We also measured case-sensitive detokenized BLEU with SacreBLEU in these experiments.

\subsubsection{Methods}
We compare our proposed strategies with baselines used in Section \ref{sec:exp_standard_setting}.
We used the Transformer (base) setting with Admin as \textbf{Vanilla} and prepared \textbf{Universal} which is $M=1, N=6$ with large dimension sizes for each internal layer.
For the proposed strategies, we used $M = 6$ and $N = 18$.
In these configurations, the training time of proposed strategies are almost the same as one of Universal as described in Table \ref{tab:exp_main_result}.

\subsubsection{Results}
Table \ref{tab:exp_other_result} shows BLEU scores of each method on each dataset.
This table indicates that Universal outperformed Vanilla in all datasets.
The proposed parameter sharing strategies (\textsc{sequence}, \textsc{cycle}, and \textsc{cycle (rev)}) achieved better scores than Universal in all datasets.
These results are consistent with results in Table \ref{tab:exp_main_result}.
These results also indicate that the proposed strategies are more efficient than Universal, which shares parameters of one layer with all layers, because they achieved better performance with almost the same parameter size and computational time.

In the comparison among the proposed strategies, \textsc{cycle} and \textsc{cycle (rev)} outperformed \textsc{sequence} on German-to-English but it is difficult to conclude that \textsc{cycle} and \textsc{cycle (rev)} are superior to \textsc{sequence} on English-to-French and French-to-English.
This result implies that the best strategy might depend on a language pair\footnote{Section \ref{sec:exp_asr} and Appendix \ref{sec:exp_lm} imply that a sort of task and Transformer architectures also have an influence on the performance of proposed strategies.}.
However, we emphasize that our proposed strategies outperformed Universal.
For applying our proposed parameter sharing strategies to other datasets, we recommend using \textsc{sequence} as a first step because it is the easiest to implement.

\section{Experiments on Automatic Speech Recognition}
\label{sec:exp_asr}

\begin{table*}[!t]
  \centering{}
  % \footnotesize
  \begin{tabular}{ l | c c c c | c | c | c c c c } \hline
         & \multicolumn{2}{c}{Enc} & \multicolumn{2}{c|}{Dec} & &  & \multicolumn{2}{c}{Dev} & \multicolumn{2}{c}{Test} \\
  Method & $M$ & $N$ & $M$ & $N$ & \#Params & Speed & clean & other & clean & other \\ \hline \hline
  % Method & \#Params & Dev (clean) & Dev (other) & Test (clean) & Test (other) \\ \hline \hline
  Vanilla & 6  & 6 & 6 & 6 & 52M & $\times$2.94& 3.98 & 9.06 & 4.18 & 9.18 \\
  Universal & 1 & 6 & 1 & 6 & 54M & $\times$1.00 & 3.73 & 8.85 & 4.14 & 8.80 \\ 
  \textsc{sequence} & 8 & 16 & 4 & 8 & 50M & $\times$1.41 & 3.16 & \textbf{7.84} & \textbf{3.32} & \textbf{7.71}\\
  \textsc{cycle} & 8 & 16 & 4 & 8 & 50M & $\times$1.41 & 3.28 & 7.86 & 3.57 & 7.97 \\
  \textsc{cycle (rev)} & 8 & 16 & 4 & 8 & 50M & $\times$1.41 & \textbf{3.11} & 8.10 & 3.60 & 8.11\\ \hline
  \end{tabular}
  \caption{The parameter sizes, computational speeds based on the Universal configuration, and word error rates of each method. For word error rates, lower is better. Scores in bold denote the best results for each set.\label{tab:exp_asr}}
\end{table*}

\subsection{Datasets}
To investigate the effect of our proposed strategies on other modality, we conduct comparisons on the automatic speech recognition (ASR) task.
We used the de-facto standard English ASR benchmark dataset: LibriSpeech~\cite{7178964}.
The dataset contains 1,000 hours of English speech from audiobooks.
We used the standard splits of LibriSpeech; used all available training data for training and two configurations (clean and other) of development and test sets for evaluation.
We applied the same pre-processing as in \cite{wang-etal-2020-fairseq}.
We measured word error rate on each set.

\subsection{Methods}
We also compare our proposed strategies with baselines in Section \ref{sec:exp_mt}.
As the base architecture, we used Transformer based speech-to-text model (T-Md) described in \cite{wang-etal-2020-fairseq}.
In contrast to the Post-LN architecture, which is the original Transformer architecture~\cite{NIPS2017_7181}, the Transformer in T-Md consists of the Pre-LN configuration.
We prepared 6 layers for the encoder and decoder in \textbf{Vanilla} and \textbf{Universal}.
For proposed strategies, we stacked more layers for the encoder side in the same as in \cite{wang-etal-2020-fairseq}.
We prepared $N=16$ and $M=8$ for the encoder side, and $N=8$ and $M=4$ for the decoder side.

\subsection{Results}
Table \ref{tab:exp_asr} shows word error rates of each method on each dataset.
This table indicates that Universal outperformed Vanilla in all sets.
The proposed parameter sharing strategies (\textsc{sequence}, \textsc{cycle}, and \textsc{cycle (rev)}) achieved better scores than Universal in all sets even though they are faster than Universal.
These results are consistent with results in machine translation experiments in Section \ref{sec:exp_mt}.
Thus, the proposed strategies are also more efficient in the ASR task.

In contrast to machine translation experiments, \textsc{sequence} outperformed \textsc{cycle} and \textsc{cycle (rev)} in the ASR task.
We consider that this result might be caused by the difference of tasks.
In addition, the cause might be the difference of layer normalization positions in the Transformer architecture.
We used Post-LN based method (Admin)~\cite{liu-etal-2020-understanding} in machine translation experiments, but Pre-LN based method in this ASR task.
\newcite{liu-etal-2020-understanding} and \newcite{takase:2022:b2t} demonstrated that the position of the layer normalization has a strong effect on the property of Transformers.
The experimental results in language modeling (Appendix \ref{sec:exp_lm}) also imply that \textsc{sequence} is more appropriate when we use the Pre-LN based Transformer.
The main focus of this study is empirical comparisons to the widely used parameter sharing strategy, Universal~\cite{dehghani2019}, but we will address theoretical analyses on the training dynamics in the future to understand the relation between parameter sharing strategies and Transformer architectures.

\section{Related Work}
\label{sec:relatedwork}
\paragraph{Parameter Sharing}
In the past decade, various studies reported that a large amount of training data improve the performance in NLP tasks~\cite{suzuki-isozaki-2008-semi,brants-etal-2007-large,NIPS2013_9aa42b31,sennrich:2016:backtrans,edunov-etal-2018-understanding}.
Moreover, recent studies indicated that the larger parameter size we prepare, the better performance the model achieves when we have a large amount of training data~\cite{devlin-etal-2019-bert,NEURIPS2020_1457c0d6}.
In fact, the best system on the WMT 2020 news translation task is composed of about 10 times as many parameters as the widely used Transformer (base) setting~\cite{kiyono-etal-2020-tohoku}.
However, due to the limitation on a GPU memory capacity, we have to explore a parameter efficient way, which achieves better performance while saving the parameter size.

Parameter sharing is a widely used technique as a parameter efficient way~\cite{dehghani2019,Dabre_Fujita_2019,Xia_He_Tan_Tian_He_Qin_2019,lan2020}.
\newcite{dehghani2019} proposed Universal Transformer.
Their method requires parameters for only one layer (i.e., $M=1$) of a Transformer-based encoder-decoder, and shares these parameters with $N$ layers.
\newcite{Dabre_Fujita_2019} investigated the effectiveness of Transformer sharing parameters of one layer across all layers on various translation datasets.
\newcite{lan2020} used this parameter sharing strategy to construct a parameter efficient model.
As reported in these studies, we can achieve better performance by the Transformer sharing parameters of one layer across all layers when we use the same parameter size as the original Transformer.
However, this strategy requires much more computational time as described in Table \ref{tab:exp_main_result} because weight matrices for each layer are much larger.
To solve this problem, we propose a new parameter sharing strategies that prepare parameters for $M$ layers and assign them into $N$ layers, where $1 \leq M \leq N$.
Experimental results show that our proposed strategies are more efficient than the method sharing parameters of one layer with across layers~\cite{dehghani2019,Dabre_Fujita_2019,lan2020}.
In addition, experimental results imply that the proposed parameter sharing strategies are effective to improve the performance.
In fact, in language modeling, previous studies demonstrated that the parameter sharing is useful to improve the performance~\cite{DBLP:journals/corr/MelisDB17,merityRegOpt,takase-etal-2018-direct},

\newcite{Xia_He_Tan_Tian_He_Qin_2019} proposed an encoder-decoder which shares parameters of the encoder part and decoder part.
\newcite{ijcai2019-735} proposed the method to share the attention weights to make the computation of Transformers fast.
These techniques are orthogonal to our proposed method.
Thus, we can combine them to improve the efficiency of parameters and computational time.

\paragraph{Training Acceleration}
In this study, we explore a parameter efficient method.
On the other hand, recent studies proposed method to accelerate the training.
\newcite{li-etal-2020-shallow} proposed a training strategy for a deep Transformer.
Their strategy trains a shallow model and then stacks layers to construct a deep model.
They repeat this procedure until the desired deep model.
They indicated that their strategy was faster than the training of whole parameters of a deep Transformer.
\newcite{takase-kiyono-2021-rethinking} compared regularization methods in terms of training time.
Their experimental results show that the simple regularizations such as word dropout are more efficient than complex ones such as adversarial perturbations.
We can use those findings to accelerate the training of our proposed strategies.

\paragraph{Deep Transformers}
To raise expressiveness power of Transformers, we stack many layers in the proposed method.
The stability of training deep Transformers depends on their architectures~\cite{iwslt-2019-transformer,layer-normalization-icml2020,liu-etal-2020-understanding}.
Transformer architectures can be categorized into two types based on the position of layer normalizations: Post-LN and Pre-LN.
Most of recent studies used the Pre-LN setting when they stacked many layers~\cite{wang-etal-2019-learning,NEURIPS2020_1457c0d6} because Pre-LN makes the training process more stable than the Post-LN setting, which is used in the original Transformer~\cite{iwslt-2019-transformer,layer-normalization-icml2020}.
On the other hand, several studies proposed methods to stabilize the training of Post-LN based Transformers~\cite{liu-etal-2020-understanding,takase:2022:b2t}.
In this study, we used Admin~\cite{liu-etal-2020-understanding} in machine translation experiments because it stabilizes the training of Post-LN based Transformers while keeping the advantages of Post-LN in the machine translation task.
For other experiments, we used the Pre-LN configuration based on the implementations of baselines.
These experiments show that our proposed strategies are effective in major two architectures: Post-LN and Pre-LN.

\section{Conclusion}
We proposed three parameter sharing strategies: \textsc{sequence}, \textsc{cycle}, and \textsc{cycle (rev)}, for the internal layers in Transformers.
In contrast to the previous strategy, which prepares parameters for only one layer and shares them across layers such as Universal Transformers~\cite{dehghani2019}, the proposed strategies prepare parameters for $M$ layers to construct $N$ layers.
The proposed strategies stack layers whose weight matrices are smaller than ones of Universal Transformers to raise expressiveness power while saving computational time.

Experimental results in the standard machine translation setting show that the proposed strategies achieved slightly better BLEU scores to those of Universal with a small computational time when we prepared almost the same parameters for each method ($M=6$ and $N=12$).
In addition, the proposed strategies outperformed Universal under the same computational budgets ($M=6$ and $N=18$).
Thus, the proposed strategies are efficient in terms of the parameter size and computational time.
Through additional experiments, we indicated that the proposed strategies are also more efficient than Universal in the high resource setting, other language pairs, and another modality (speech-to-text).

\section*{Limitations}
As described in Section \ref{sec:intro}, the purpose of this study is to relax the existing parameter sharing strategy which shares the parameters of one layer with all layers~\cite{dehghani2019,Dabre_Fujita_2019,lan2020}.
Experimental results indicate that the proposed simple parameter sharing strategies can be a better alternative to the existing method.
As many studies on neural methods, this study also depend on empirical observations.
In other words, this study lacks theoretical justifications for proposed parameter sharing strategies.

We conducted experiments on various situations.
We mainly focused on sequence-to-sequence tasks and trained each model from scratch.
Our conducted experiments indicated the efficiency of the proposed strategies but we did not conduct experiments on the pre-training and then fine-tuning configuration such as comparison with BERT~\cite{devlin-etal-2019-bert} due to the limitation of our computational budgets.
Thus, it is difficult to claim that the proposed strategies are also more efficient in such configuration.
In addition, we have to investigate the effectiveness in a more realistic situation.
For example, we will investigate the performance of the combination of our proposed method, which is the parameter efficient way for internal layers, and a parameter efficient embedding such as \newcite{takase2020word}.

Through experiments in various configurations, it is difficult to conclude which strategy is the best.
Experimental results imply that the best strategy depends on the task and Transformer architecture (Post-LN or Pre-LN).
Such phenomena are reported in previous studies~\cite{press-etal-2020-improving,gulati2020conformer}.
In fact, the architecture explored by \newcite{press-etal-2020-improving} is better in the language modeling task but ineffective in the machine translation task.
Since it is intractable to investigate a tremendous amount of possible parameter assignment way due to the limitation of computational budgets, there might be a superior way to three simple strategies proposed in this paper.
However, we emphasize that all our proposed strategies are more efficient than the Universal configuration.
Because the purpose of our experiments is not to detect the best parameter sharing strategy but to indicate that our proposed parameter sharing strategies are more efficient than the Universal configuration, we consider that our conducted experiments are sufficient to verify our claims.

\section*{Ethics Statement}
As discussed in \citet{strubell-etal-2019-energy}, recent neural models require substantial energy consumption.
To address this issue, we explore a parameter efficient way for Transformers in this study.
We believe that our proposed strategies are effective to reduce the energy consumption.

On the other hand, we spent a large amount of computational costs to investigate the usefulness of our proposed strategies in various situations.
Appendix \ref{appendix:hyperparameter} indicates our used GPUs and the number of updates that correspond to the computational costs.

\section*{Acknowledgements}
We thank the anonymous reviewers for their insightful suggestions.
A part of this work was supported by JSPS KAKENHI Grant Number JP21K17800 and JST ACT-X Grant Number JPMJAX200I.

\bibliographystyle{acl_natbib}
% \bibliography{reference}

\clearpage

\appendix

\section{Experiments on Language Modeling}
\label{sec:exp_lm}

\subsection{Dataset}
We focused Transformer-based encoder-decoders in the main experiments of this paper.
However, recent studies often employed the decoder side only as a pre-trained model.
Thus, we conduct experiments on the language modeling task to investigate the efficiency of our proposed strategies when we use the decoder side only.
We used Wikitext-103~\cite{DBLP:journals/corr/MerityXBS16} which contains a large amount of training data.
We measured perplexity of validation and test sets.

\subsection{Methods}
We used the Transformer with adaptive inputs~\cite{DBLP:journals/corr/abs-1809-10853} as the base architecture.
In the same as in \newcite{DBLP:journals/corr/abs-1809-10853}, the Transformer in the language modeling consists of the Pre-LN configuration.
We set $N=6$ for \textbf{Vanilla} and \textbf{Universal}.
For the proposed strategies, we set $N=12$ and $M=6$.

\subsection{Results}
Table \ref{tab:exp_lm} shows perplexities of each method.
This table indicates that Vanilla achieved better performance than Universal.
Thus, the sharing parameters of one layer with all layers might not be suitable for a large-scaled language modeling task.
In contrast, the proposed strategies outperformed Vanilla.
This result indicates that our proposed strategies are also more efficient than Universal in the language modeling.

Through the comparison among proposed strategies, \textsc{sequence} achieved the best perplexity.
As described in Section \ref{sec:exp_asr}, \textsc{sequence} might be more appropriate to the Transformer with the Pre-LN configuration.
To explore the reason, we believe that we have to conduct the theoretical analysis of the Transformer during its training.
We address this issue in the future study.

The lower part of Table \ref{tab:exp_lm} shows the reported score of \newcite{DBLP:journals/corr/abs-1809-10853}, our reproduced score, and \textsc{sequence} with more parameters.
This part indicates that \textsc{sequence} achieved better perplexities than others even though the parameter size of \textsc{sequence} is smaller.
Therefore, \textsc{sequence} is also efficient when we prepare a large amount of parameters for a language model.

\begin{table}[!t]
  \centering{}
  \footnotesize
  \begin{tabular}{ l | c | c c } \hline
  Method & \#Params & Valid & Test \\ \hline \hline
  Vanilla & 121M & 20.39 & 21.13 \\
  Universal & 121M& 22.75 & 23.84 \\
  \textsc{sequence} & 121M & \textbf{18.97} & \textbf{19.69} \\
  \textsc{cycle} & 121M & 19.00 & \textbf{19.69} \\
  \textsc{cycle (rev)} & 121M & 19.60 & 20.24 \\ \hline \hline
  \multicolumn{4}{c}{Models with more parameters} \\ \hline \hline
  \newcite{DBLP:journals/corr/abs-1809-10853}$\dagger$ & 247M & 18.53 & 19.24 \\
  \newcite{DBLP:journals/corr/abs-1809-10853} & 247M & - & 18.7 \ \ \\
  \textsc{sequence} & 234M & \textbf{17.71} & \textbf{18.55} \\ \hline
  \end{tabular}
  \caption{The parameter sizes and perplexities of each method. The lower part indicates scores reported in \newcite{DBLP:journals/corr/abs-1809-10853} and the score of \textsc{sequence} with more parameters. Scores in bold denote the best results for each set. $\dagger$ represents our re-run of \citet{DBLP:journals/corr/abs-1809-10853}. \label{tab:exp_lm}}
\end{table}

\section{Details of Experimental Settings}
\label{appendix:hyperparameter}
We used NVIDIATesla V100 GPUs for all experiments.
Table \ref{tab:hyp_params} shows the hyper-parameters for training in each task.
The descriptions in our code also help to understand configurations in this study.
\begin{table*}[!t]
  \centering
  % \vskip -0.2in
  % \footnotesize
  % \vskip 0.1in
  \begin{tabular}{ l | c c c} \hline
  Params & Machine Translation & ASR & Language Model \\ \hline
  Leaning rate & 0.001 & 0.001 & 0.001 \\
  Scheduler & inverse sqrt & inverse sqrt & inverse sqrt \\
  Adam \ $\beta$ & (0.9, 0.98) & (0.9, 0.98) & (0.9, 0.98) \\
  Warmup updates & 4k & 4k & 2k \\
  Max updates & 50k & 150k & 50k \\ \hline
  \end{tabular}
  \caption{Hyper-parameters used in our experiments.\label{tab:hyp_params}}
\end{table*}

\end{document}